%% file: main.tex
\documentclass[conference]{IEEEtran}
\usepackage{times}

\usepackage[numbers]{natbib}
\usepackage{multicol}
\usepackage[bookmarks=true]{hyperref}
\usepackage[utf8]{inputenc}
\usepackage[T1]{fontenc}
\usepackage{url}
\usepackage{booktabs}
\usepackage{amsfonts}
\usepackage{nicefrac}
\usepackage{microtype}
\usepackage[pdftex]{graphicx}
\usepackage{amsmath}
\usepackage{xcolor}
\usepackage{textcomp}

\DeclareRobustCommand*{\IEEEauthorrefmark}[1]{%
  \raisebox{0pt}[0pt][0pt]{\textsuperscript{\footnotesize #1}}%
}

\let\labelindent\relax
\usepackage{enumitem}

\title{Learning Gentle Object Manipulation with Curiosity-Driven Deep Reinforcement Learning}

\author{
    \IEEEauthorblockN{Sandy H. Huang*\IEEEauthorrefmark{1}, Martina Zambelli*\IEEEauthorrefmark{2}, Jackie Kay\IEEEauthorrefmark{2}, Murilo F. Martins\IEEEauthorrefmark{2},\\ Yuval Tassa\IEEEauthorrefmark{2}, Patrick M. Pilarski\IEEEauthorrefmark{2}, Raia Hadsell\IEEEauthorrefmark{2} }
    \IEEEauthorblockA{\IEEEauthorrefmark{1}University of California, Berkeley}
    \IEEEauthorblockA{\IEEEauthorrefmark{2}DeepMind}\\\vspace{-0.5cm}
    \small\texttt{shhuang@cs.berkeley.edu}, \texttt{\{zambellim,tassa,kayj,murilomartins,ppilarski,raia\}@google.com}
}

\newcommand{\sref}[1]{Sec. \ref{#1}}
\newcommand{\figref}[1]{Fig. \ref{#1}}

\newcommand\crule[3][black]{\textcolor{#1}{\rule{#2}{#3}}}
\definecolor{plot-r}{HTML}{3b82ba}  % blue
\definecolor{plot-r-rg}{HTML}{999999}  % gray
\definecolor{plot-r-rg-rs}{HTML}{d99e38}  % yellow
\definecolor{plot-r-rg-rsp1}{HTML}{8c82d1}  % purple
\definecolor{plot-r-rg-rsp2}{HTML}{8cab42}  % green
\definecolor{plot-r-rg-rsp3}{HTML}{963333}  % red
\definecolor{plot-r-rg-rsp4}{HTML}{42ab60}  % green real robot
\definecolor{plot-r-rg-rsp5}{HTML}{cf83b9}  % pink real robotr-rg-rs
\definecolor{plot-r-rg-rsp6}{HTML}{ffff00}  % light yellow
\definecolor{plot-r-rg-rsp7}{HTML}{f19d39}  % light orange

\begin{document}

\maketitle

\input{abstract.tex}
\input{introduction.tex}
\input{related_work.tex}
\input{problem_statement.tex}
\input{approach.tex}
\input{experiments.tex}
\input{discussion.tex}

\bibliographystyle{plainnat}
\bibliography{references}

\end{document}

%% file: abstract.tex
\begin{abstract}
Robots must know how to be \emph{gentle} when they need to interact with fragile objects, or when the robot itself is prone to wear and tear. We propose an approach that enables deep reinforcement learning to train policies that are gentle, both during exploration and task execution. In a reward-based learning environment, a natural approach involves augmenting the (task) reward with a penalty for non-gentleness, which can be defined as excessive impact force. However, augmenting with only this penalty impairs learning: policies get stuck in a local optimum which avoids all contact with the environment. Prior research has shown that combining auxiliary tasks or intrinsic rewards can be beneficial for stabilizing and accelerating learning in sparse-reward domains, and indeed we find that introducing a surprise-based intrinsic reward does avoid the no-contact failure case. However, we show that a simple dynamics-based surprise is not as effective as \emph{penalty-based surprise}. Penalty-based surprise, based on predicting forceful contacts, has a further benefit: it encourages exploration which is contact-rich yet gentle. We demonstrate the effectiveness of the approach using a complex, tendon-powered robot hand with tactile sensors. Videos are available at \url{http://sites.google.com/view/gentlemanipulation}.
\end{abstract}

%% file: introduction.tex
\section{Introduction}
Deep reinforcement learning (RL) can be used to train policies that achieve superhuman performance on Atari games~\cite{Mnih_2013} and Go~\cite{Silver_2016}, learn locomotion tasks \cite{Lillicrap_2016,Schulman_2015}, and perform complex robotic manipulation skills~\cite{Levine_2016}. However, deploying deep RL on real-world robots often leads to a considerable amount of wear and tear over time, on both the robot itself and the environment, because existing approaches require many trials to learn and often rely on simple stochastic exploration. If robots were able to explore and learn safely, minimizing excessive forces and impacts, they would last longer before needing repairs, and the objects they interact with would not need to be replaced as often. 

Moreover, destructive or risky behaviors by robots trained with RL go beyond the exploration phase. Agents trained solely to maximize a reward signal will often converge on policies which are high velocity and thus high impact, resulting in so-called ``bang-bang control,'' which may be optimal in terms of returns, but is potentially damaging or dangerous to the robot and the environment. 
As a further motivation, we might care about gentleness in terms of task execution itself, for instance if the robot needs to pick-and-place an object, but either the object and/or goal location is fragile. In particular, when robot manipulation involves humans (e.g., feeding a patient), being gentle is important~\cite{Ikuta_2003}. In these situations, we would like robots to accomplish the given task in a reasonable amount of time, while minimizing applied force and impact as much as possible. 

Thus, in order to broadly deploy deep RL on real robots, we need an approach for training policies that are gentle, both during exploration and task execution. A na\"ive approach is to constrain the maximum torques that a robot's motors can exert. However, many manipulation tasks require occasional, momentary, or variable high force (e.g., hammering a nail or turning a lever); the torque limit cannot be any lower than this, otherwise the robot will not be able to complete the task. But we do not want the robot to freely exert this much force along its entire trajectory. Alternatively, one could constrain the total amount of force or impact allowed, but this requires knowing \emph{a priori} the minimum total amount necessary for accomplishing the task~\cite{Achiam_2017,Altman_1999,Tessler_2018}.

Instead, our approach is to give the robot negative rewards for actions that are not gentle, for instance those that result in high impact forces. Incorporating this in the reward function is a natural approach for encoding preferences about \emph{how} robots should perform a task (e.g., driving style~\cite{Abbeel_2004}), and can be seen as an intrinsic ``pain'' signal that encourages learning safer policies. However, perhaps unsurprisingly, adding this penalty often makes it much harder for an agent to learn a successful policy; instead, it gets stuck in a local optimum of avoiding contact altogether, because it encounters the penalties before ever obtaining the task reward, and thus learns a policy that is dominated by aversion to pain.

To motivate agents to interact with the environment \emph{and} do it gently, we propose balancing this ``pain'' signal by adding another intrinsic signal, this one positive, for curiosity. In particular, we reward the agent for \emph{surprising} experiences---those that contradict the agent's current understanding of the world. A concrete example of this is giving intrinsic rewards for transitions that have low probability under a learned dynamics model~\cite{Achiam_2017b,Houthooft_2016,Pathak_2017,Stadie_2015}. However, we find that using this kind of \emph{dynamics-based surprise} is not as effective as using a \emph{penalty-based surprise}, that leads robots to be explicitly curious about the non-gentleness penalty itself. In this formulation, the agent makes predictions about the pain penalty that will result from a given state and action, and erroneous predictions deliver a small positive reward. 

Although this curiosity about pain may seem somewhat counterintuitive, it has been shown that humans are specifically curious about painful or unpleasant experiences~\cite{hsee2016curiosity}; moreover, it is well known that children engage in physical risk-seeking behavior~\cite{Morrongiello_2007,Sandseter_2007,Scheidt_1995}. This is also observed in other species in the form of play fighting~\cite{Spinka_2001}, and seems to have an evolutionary benefit. Research in developmental psychology suggests that risky play is essential to the development of children, by allowing them to test their physical limits, improve hand-eye coordination, and learn to avoid or adapt in dangerous environments~\cite{Brussoni_2012,Jambor_1986}.

Motivated by this specialized form of curiosity, our work takes a step toward using deep RL to train policies for gentle, object- and contact-centric manipulation. In this work, gentleness is defined as minimizing impact forces. We demonstrate that our proposed approach, which introduces both a penalty for excessive impact forces and a curiosity reward focused on this penalty, enables efficient and safe exploration, precise task execution, and successful manipulation of fragile objects.

%% file: related_work.tex
\section{Related Work}
Our goal of gentle manipulation is closely related to impact minimization in classical control, but existing approaches typically rely on having accurate dynamical contact models \cite{Hu_2017, Huang_2006, Wee_1993}. Another related domain is safe reinforcement learning \cite{Garcia_2015, Pecka_2014}: safety may refer to either physical safety or handling environment stochasticity. Typically, the former is concerned with avoiding catastrophic situations (e.g., crashing into another car or falling off a cliff), whereas in our work, we take a broader view of physical safety, in terms of reducing wear-and-tear in order to delay \emph{eventual} damage.

In the context of developmental robotics, curiosity and intrinsic motivation take inspiration from developmental psychology, where they are believed to be essential for achieving autonomous mental development. This is realized by using an intrinsic curiosity drive that encourages a robot to focus on situations that are neither too simple (or predictable) nor too hard (or unpredictable) \cite{baranes2014effects, oudeyer2009intrinsic, oudeyer2007intrinsic, oudeyer2013intrinsically}. These approaches, which often rely on engineered features and action scripts or primitives, have claimed that curiosity can drive a self-guided curriculum to train a robot arm for block manipulation and stacking~\cite{ngokatanarobot2012}. Our approach follows in this vein, using curiosity-based intrinsic rewards to encourage agents to (gently) explore their environment, in the presence of impact penalties. Curiosity-based intrinsic rewards can also be viewed as providing reward shaping \cite{Ng_1999}: they make tasks easier to learn, by making it less likely for agents to get stuck in undesirable local optima. 

In the context of deep RL, curiosity-based intrinsic rewards typically reward agents for either encountering novel states \cite{Bellemare_2016,Fu_2017,Tang_2017} or encountering surprising experiences \cite{Achiam_2017b,Houthooft_2016,Pathak_2017,Stadie_2015}. Recently, researchers have shown that intrinsic motivation can be used to learn human-like social skills in a human-robot interaction domain~\cite{qureshinn2018}. Our work investigates surprise-based intrinsic rewards, but we find that the usual method of computing this with respect to a learned dynamics model is not effective in our setting. Instead, we formulate surprise with respect to a penalty-prediction model.

In our work, the reward comes from a combination of several signals: extrinsic rewards from the task (positive), and intrinsic rewards from impacts (negative) and curiosity (positive). This falls under multi-objective reinforcement learning (MORL). Typically MORL approaches either train a single policy by finding the right balance of rewards, or learn a set of policies that approximate the Pareto optimal frontier \cite{Liu_2015, Roijers_2013}. Some authors have used the Constrained MDP framework \cite{Altman_1999} to develop policy optimization methods which ensure that constraints are satisfied at all points throughout learning \cite{Achiam_2017,dalal2018safe}. 
While such methods could potentially augment our approach, we believe a simpler mechanism might suffice. We assume the balance between ``pain'' and task achievement should be encoded in the reward structure and naturally found by the agent; thus if it is very important to achieve the task, then a higher task reward conveys that being less gentle (e.g., experiencing higher impact forces) is acceptable. From this standpoint, we directly add these rewards together, rather than searching for a weighting that leads to the correct behavior.

An important aspect of our method is learning with tactile sensors.
In \cite{amos2018learning} touch was used, along with proprioception and vestibular information, to learn awareness models that predict proprioceptive information about the agent's body and also represent objects in the external world.
Touch has also been integrated in robotic exploration and model learning; it has been shown to improve performance in discriminating objects and in solving tasks involving multiple sensory modalities \cite{amos2018learning, kaboli2017tactile, kroemer2011learning, schmitz2014tactile, zambelli2016multimodal}. However, these latter approaches do not leverage intrinsic motivation to improve exploration strategies.

%% file: problem_statement.tex
\section{Preliminaries}
\subsection{Markov Decision Process}
A Markov Decision Process (MDP) is defined as a tuple $(\mathcal{S}, \mathcal{A}, \mathcal{P}, \mathcal{R}, \gamma)$, where $\mathcal{S}$ is the state space, $\mathcal{A}$ is the action space, $\mathcal{P}: S \times A \times S \mapsto \mathbb{R}$ specifies the transition probabilities, $\mathcal{R}: S \times A \times S \mapsto \mathbb{R}$ specifies the reward function, and $\gamma \in [0, 1]$ is the discount factor.

A policy $\pi$ is a function that maps each state to a distribution over actions ($\pi: \mathcal{S} \to \Delta_{\mathcal{A}}$, where $\Delta_{\mathcal{A}}$ is the probability simplex on $\mathcal{A}$). Reinforcement learning optimizes policies to maximize expected returns (i.e., cumulative discounted future rewards):
\begin{equation*}
    \mathbb{E}_{a_t \sim \pi(s_t), s_{t+1} \sim \mathcal{P}(s_t, a_t)}[\sum_{t} \gamma^t \mathcal{R}(s_t,a_t,s_{t+1})].
\end{equation*}

A policy $\pi$'s action-value function is
\begin{align*}
\begin{split}
    &Q^\pi(s, a) = \\
    & \;\;\;\; \int_{s'}P(s, a, s') \left(\mathcal{R}(s, a, s') + \gamma \mathbb{E}_{a' \sim \pi(s')} [Q^\pi(s', a')]\right).
\end{split}
\end{align*}

Typically $\mathcal{R}$ specifies how well the policy is doing in terms of accomplishing a task. In this work, we augment $\mathcal{R}$ with several types of auxiliary rewards, in order to train policies for contract-centric, low-impact manipulation.

\subsection{Deep Reinforcement Learning}
The policy $\pi$ can be represented by a function parameterized by $\theta$; for instance, $\theta$ may be a weighting on predefined features of the state \cite{Abbeel_2004}. In deep RL, $\theta$ is the parameterization of a neural network. We use Distributed Distributional Deterministic Policy Gradients (D4PG) \cite{Maron_2018} to train our policies, but in principle our proposed approach is algorithm-agnostic.

D4PG is an actor-critic algorithm used to train policies for continuous control, where both the actor and critic are parameterized by neural networks. The critic is a distributional action-value function: it takes the current state $s_t$ and action $a_t$ as input, and outputs a categorical distribution over the predicted $Q(s_t, a_t)$. It is trained with off-policy evaluation, on batches of transitions $(s_t, a_t, s_{t+1})$ sampled from a replay buffer. The actor is a deterministic policy: it takes in the current state $s_t$ as input, and outputs an action $a_t$. During training, the actor's policy is updated using gradients that are computed only with respect to the critic, such that actions are adjusted in the direction of increased Q-values.

\subsection{Formalizing gentleness}
In this work, we define being gentle as minimizing impact. This is closely related to the notion of impact force in physics, which is the maximum amount of force experienced during a collision. However, we consider a more general definition of ``impact,'' that does not only apply to cases when the initial applied force is zero. Instead, assuming a discrete time step, we define impact $m_t$ as
\begin{equation}
    m_t = \max(0, f_{t+1} - f_t),
\end{equation}
where $f_t$ is the sensed force at time step $t$. In other words, for a robot to be gentle, it should minimize \emph{increases in sensed force}. To illustrate, consider a robot that needs to push a heavy object with a force of 20N. If the robot increases the applied force from zero to 20N in a fraction of a second, the action is more likely to cause damage compared to amortizing the increase in force over several seconds.

%% file: approach.tex
\section{Proposed Approach}

In order to train policies that exhibit gentle manipulation, we propose to augment the original reward ($r_t$) with an impact force penalty ($r_t^f$) and an intrinsic reward based on surprise ($r_t^s$). Agents are trained to maximize the total expected return,
\begin{equation}
    r_t' = r_t + r_t^f + r_t^s.
\end{equation}

\subsection{Impact penalty}
\label{sec:penalty}
The impact force penalty acts as an intrinsic pain signal to encourage agents to accomplish manipulation tasks in a more gentle way. Of course, in order to accomplish any manipulation task, small impacts are necessary---at some point the robot needs to go from zero to non-zero applied force on an object, in order to manipulate it. So, the impact penalty should scale non-linearly with the level of impact, by taking into account the \emph{acceptability} of a particular amount of impact.

Let $a_\lambda(m) \in [0, 1]$, parametrized by $\lambda$, be the acceptability of an amount of impact $m$. This is a monotonically increasing function, that should be designed according to how resilient the robot and environment are to impacts. For instance, if the robot is interacting with very fragile objects, then the range of acceptable impacts should be smaller. Ideally, this function should express the probability of damage (to either robot or environment) from a given amount of impact, and could be learned from experience of actual damage. In our experiments, since we do not have enough data on damage to estimate likelihoods, we use a sigmoid function for acceptability:
\begin{equation}
    a_\lambda(m) = \text{sigmoid}(\lambda_1 (-m + \lambda_2)) = \frac{1}{1 + e^{\lambda_1 (m - \lambda_2)}}
\end{equation}
The impact penalty at time step $t$ is then:
\begin{equation}
    r^f_t = -\sum_i (1-a_\lambda(m^i_t))  m^i_t,
\end{equation}
where the sum is over force sensors at different locations on the robot (e.g., the fingers of a robot hand). In our experiments, we set $\lambda = [2,2]^\top$.

However, if the environment reward merely combines the task reward and the impact penalty, that is, $r_t' = r_t + r_t^f$, we find that policies get reliably stuck in a local optimum of not making contact with anything in the environment---the agent learns to be afraid of contact, since it encounters the impact penalty before the sparse task reward, hindering exploration.

\subsection{Dynamics-based surprise}
The purpose of adding surprise-based intrinsic rewards is to encourage policies to make \emph{contact} with objects in the environment but still in a \emph{gentle} way. For an agent to be ``surprised,'' it must have some predictor of future states, i.e., a model. In the case of dynamics-based surprise, this model is a learned dynamics model that takes in the current state and action, and predicts the mean and variance of the next state. We train an ensemble of neural networks for the dynamics model, in order to have predictive uncertainty \cite{Balaji_2017}. Predictive uncertainty is useful for capturing novelty: in the case of environments with deterministic dynamics, if the networks in the ensemble either individually have high variance in their predictions, or have high variance across the ensemble, then this indicates a novel area that should be explored further.

Each of the $M$ networks in the ensemble outputs the mean and variance of a Gaussian for each dimension $d$ of the prediction. The ensemble's combined output is a mixture of Gaussians for each output dimension $d$: $$\frac{1}{M} \sum_i \mathcal{N}(\mu_{\theta_i}(\mathbf{x})_d, \sigma^2_{\theta_i}(\mathbf{x})_d), $$ where $\mathbf{x}$ denotes the input and $\theta_i$ are the parameters of the $i$th network in the ensemble.
During training, each network is randomly initialized, and they are trained on different batches of transitions. We choose $M = 5$, as recommended by related work \cite{Balaji_2017}.

To compute dynamics-based surprise intrinsic reward $r_t^s$, we approximate the dynamics model's predicted distribution over next states with a single Gaussian per output dimension $d$, to measure how much variance there is \emph{across} networks in the ensemble. The mean and variance of this is
\begin{align*}
    & \mu_*(\mathbf{x})_d = \frac{1}{M} \sum_i \mu_{\theta_i}(\mathbf{x})_d \\
    & \sigma^2_*(\mathbf{x})_d = \frac{1}{M} \sum_i (\sigma^2_{\theta_i}(\mathbf{x})_d + \mu^2_{\theta_i}(\mathbf{x})_d) - \mu_*^2(\mathbf{x})_d.
\end{align*}
Then the intrinsic reward is the negative log-likelihood of the true next state under this predicted distribution over next states:
\begin{equation}
    r_t^s = -\sum_d \log \, \mathcal{N}(s_{t+1,d} \, | \, \mu_*(s_t, a_t)_d, \sigma^2_*(s_t, a_t)_d).
\end{equation}
This intrinsic reward is computed with respect to a target dynamics model, which is updated every 5000 iterations; this makes training more stable, so that the agent is not trying to surprise a model that is constantly changing. In addition, we wait for the dynamics model to become more accurate before providing intrinsic rewards to the agent: after 20,000 training steps for experiments in simulation, and after 8,000 training steps on the real robot.

\begin{figure}[!ht]
\centering
\includegraphics[width=0.98\linewidth]{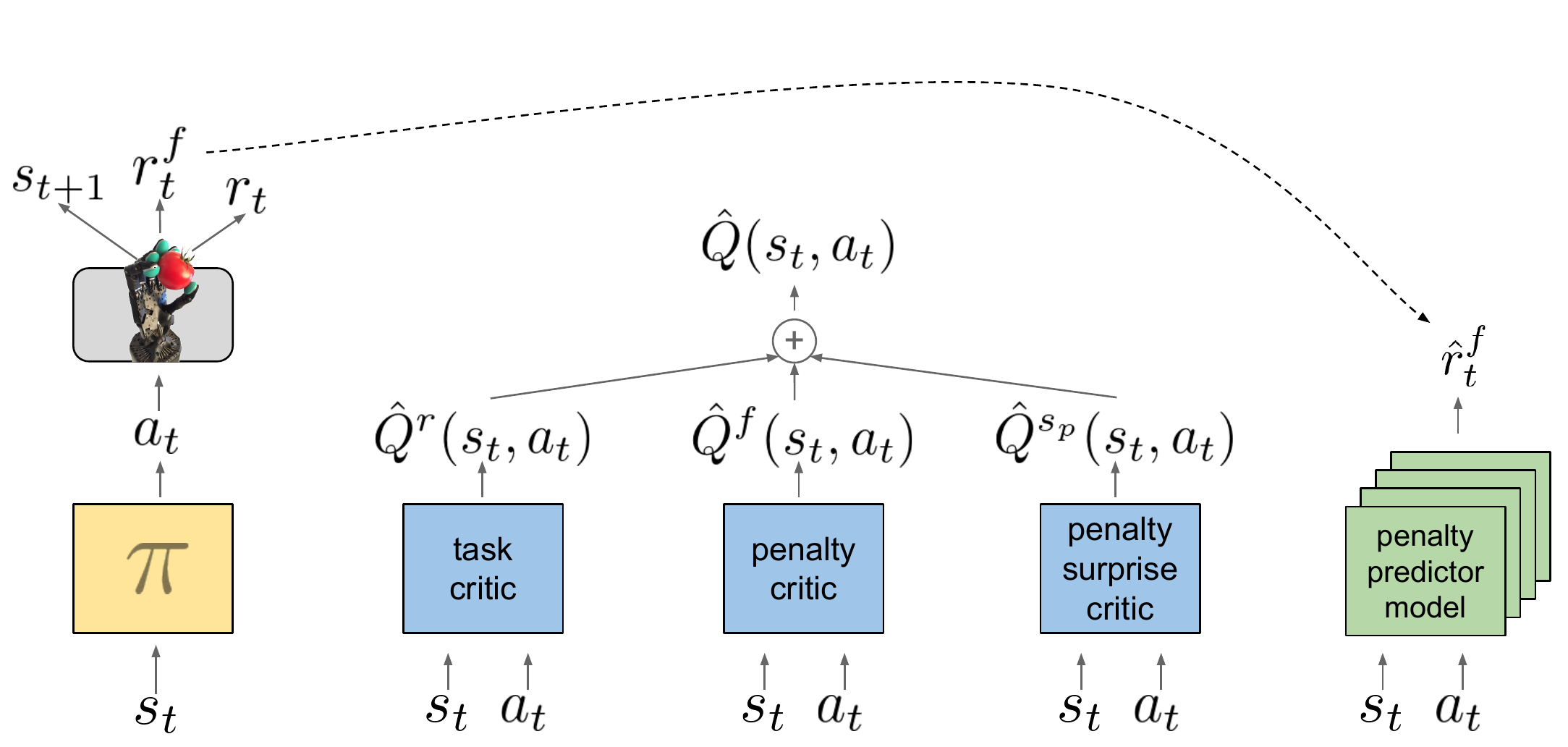}
\caption{Illustration of the architecture of the agent which uses penalty-based surprise. The policy, shown in yellow, takes actions on the environment, producing the next state, the task reward, and the penalty. There are three identical critics (a task reward critic, a penalty-surprise critic, and a penalty critic), in blue, which estimate action-values with respect to the different reward components. The penalty surprise reward is derived from the prediction error of the penalty predictor model, which is an ensemble of (five) differently-seeded networks.}
\label{fig:arch}
\end{figure}

\subsection{Penalty-based surprise}
Motivated by the role of risk-seeking behavior in childhood learning, we propose to reward the agent for being curious about the impact penalty itself. In other words, we add a reward to focus the learning and exploration of the agent on the intrinsic pain signal, with two motivations: to enable better prediction of pain, and to encourage contacts by mitigating some of the penalty. To compute the penalty-based surprise reward $r_t^{s_p}$, we train an impact penalty predictor in parallel with the agent, with the same implementation as the general dynamics model (an ensemble of five neural networks).

To compute the intrinsic reward, we do not use the negative log-likelihood directly, as was done with the dynamics reward, because it would make high impact forces acceptable as long as the prediction likelihood in those areas was low enough, leading to very non-gentle actions by the agent. Rather, we would like agents to focus on learning about areas with low penalty (i.e., areas where impacts occur but are small), so that they learn how to be gentle. For areas of high penalty, it is enough for the agent to just know that the penalty is high, not necessarily \emph{exactly} how high it is. To enforce this preference for exploring areas with low penalty while avoiding ones with high penalty, a natural approach is to augment the task reward with a convex combination between the negative log-likelihood and the impact penalty:
\begin{align}
\begin{split}
    & a_{\lambda'}(r^f_t) \, * \, -\log \, \mathcal{N}(r^f_t \, | \, \mu_*(s_t, a_t), \sigma^2_*(s_t, a_t))) \\
    & + \, (1 - a_{\lambda'}(r^f_t)) \, * \, r_t^f,
    \label{eqn:convexcomb}
\end{split}
\end{align}
where $a_{\lambda'}(r_t^f) \in [0, 1]$ is the acceptability of a particular penalty $r_t^f$. This is a monotonically increasing function, that should be chosen based on how much penalty the robot may experience for the sake of exploration or task completion. We use a sigmoid function for this acceptability, as we did for impact $a_\lambda(m)$ (\sref{sec:penalty}).

Note that $\lambda$ and $\lambda'$ play different roles: $\lambda$ modulates the mapping of impact forces to pain penalties, whereas $\lambda'$ controls the trade-off between these pain penalties and the agent's penalty-focused curiosity. The choice of $\lambda'$ may be adjusted dynamically during learning; the higher $\lambda_2'$ is, the easier it is for the robot to learn the task, at the cost of higher impact forces on average.

In order to augment the reward with this convex combination, we need to choose $r_t^{s_p}$ such that $r_t^{s_p} + r_t^f$ is equal to \eqref{eqn:convexcomb}. In addition, we only provide this intrinsic reward if the penalty is  non-zero, because the purpose is to encourage the agent to (cautiously) learn more about the penalty. Based on this, we set the penalty-based surprise intrinsic reward to be:
\begin{equation}
    r_t^{s_p} =
    \begin{cases}
      a_{\lambda'}(r^f_t) \left[ -\log \, \mathcal{N}(r^f_t \, | \, \mu_*(s_t, a_t), \sigma^2_*(s_t, a_t)) - r_t^f \right] & \\ \text{$\quad$ if } r_t^f < 0  &\\
      0 \text{ otherwise.}
    \end{cases}
\end{equation}
In the same way as dynamics-based surprise, this penalty-based surprise intrinsic reward is computed with respect to a target impact penalty predictor model, which is updated every 1000 iterations, and we do not provide intrinsic rewards to the agent until after 20,000 training steps for simulation experiments, and after 8,000 training steps for real robot experiments.

\begin{figure*}
\centering
\includegraphics[width=0.85\linewidth]{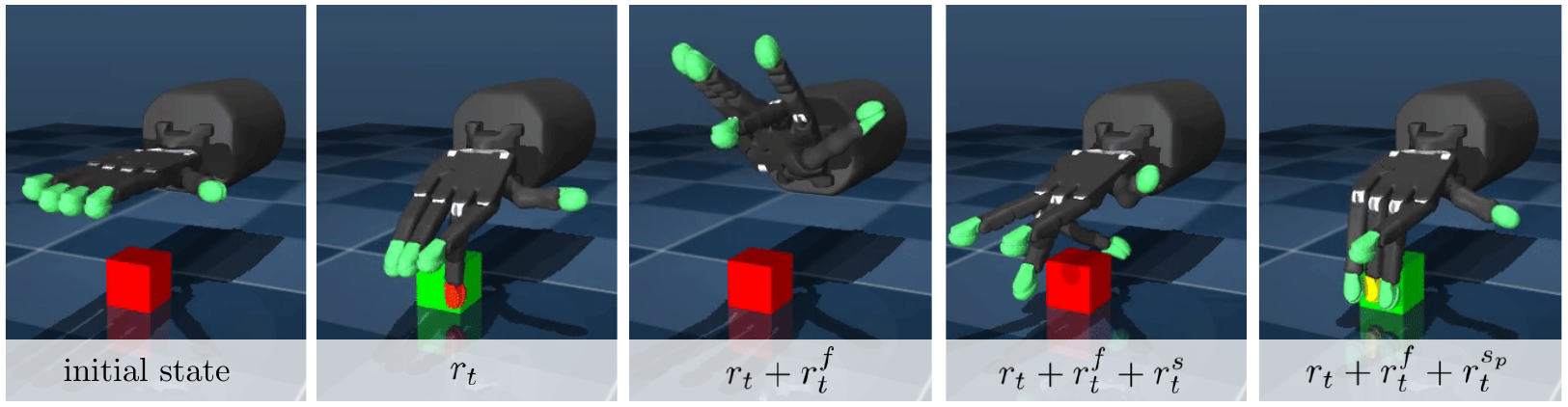}
\caption{The task is to apply at least 5N of force to the block; green block indicates success. Fingertip color shows the amount of impact force, from yellow (near-zero) to red (10N). Policies were trained for 500k iterations. Policies trained on only task reward, $r_t$, learn how to do the task, but do it a high-impact way. In contrast, policies trained on the combination of task reward, an impact penalty, and penalty-based surprise intrinsic reward ($r_t + r_t^f + r_t^{s_p}$) learn to achieve the task in a \emph{gentle} (i.e., low-impact) way, by gradually increasing force applied to the block. Without the penalty-based intrinsic reward, policies get trapped in a local optimum of avoiding contact with the environment ($r_t + r_t^f$ and $r_t + r_t^f + r_t^s$). Corresponding videos are available at \url{http://sites.google.com/view/gentlemanipulation}.
}
\label{fig:gentlemanip}
\end{figure*}

\subsection{Agent architecture and implementation details}
As mentioned, we use D4PG to train an actor (e.g., policy) and critic (e.g., action-value function). We use a separate critic for each of the reward components: the task reward critic encodes $\hat{Q}^t(s_t,a_t)$, the penalty-based surprise critic encodes $\hat{Q}^{s_p}(s_t,a_t)$, the dynamics-based surprise critic encodes $\hat{Q}^s(s_t,a_t)$, and the impact penalty critic encodes $\hat{Q}^f(s_t,a_t)$. The separation of the critics supports more stable learning \cite{Russell_2003}. The components of the agent with penalty-based surprise are illustrated in Figure~\ref{fig:arch}.

The output of the actor is passed through a $\text{tanh}$, so that it is between -1 and +1. This output specifies delta position: it is added to the current position and then clipped based on the minimum and maximum joint angle per action dimension, to obtain the action. The actor network consists of two fully-connected layers of 300 and 200 hidden units each. Each of the critic networks consists of two fully-connected layers of 400 and 300 hidden units each. The distributional output of the critic has support $(-100, 100)$ and 101 bins. For both the actor and critic networks, the first hidden layer is followed by layer normalization \cite{Ba_2016} and a $\text{tanh}$, and all other hidden layers are followed by exponential linear unit (ELU) activations. For D4PG, we used a batch size of 256 and a replay buffer of 1 million transitions.

The dynamics model consists of three ensembles, one each for predicting the three types of state features: joint position, joint velocity, and touch. The non-gentleness predictor model consists of a single ensemble. Each of these ensembles consists of five neural networks, with three fully-connected layers of 128 hidden units each. All hidden layers are followed by rectified linear unit (ReLU) activations.

%% file: experiments.tex
\section{Experiments}
Our goal is to learn policies that are safer, with less forceful impacts, while also improving sample efficiency and overall task performance. The following experiments compare three approaches for achieving this; these approaches differ in terms of what the task reward is augmented with:
\begin{itemize}[noitemsep,nolistsep]
  \item an impact penalty ($r_t^f$)
  \item an impact penalty and a dynamics-based surprise intrinsic reward ($r_t^f + r_t^s$)
  \item an impact penalty and a penalty-based surprise intrinsic reward ($r_t^f + r_t^{s_p}$).
\end{itemize}

\subsection{Experimental domain}
We run experiments both in simulation with MuJoCo~\cite{Todorov_2012} and on a physical robot. The robot platform we use is the Shadow Dexterous Hand~\cite{ShadowHand}, with five fingers and a total of 24 degrees of freedom, actuated by 20 motors. We use this platform for several reasons: because it is actuated by antagonistic tendons, it is more susceptible to wear-and-tear (and thus gentle exploration and manipulation has greater potential benefit); it can be equipped with high-fidelity tactile sensors; and it is anthropomorphic and well suited for handling fragile objects.

In simulation, each fingertip has a spatial touch sensor attached, with three channels and a spatial resolution of $4\times4$: one for normal force and two for tangential forces. We simplify this by taking the absolute value and then summing across the spatial dimensions, to obtain a 3D force vector for each fingertip. The impact force $m_t^i$ is then the sum over the increase in force per channel for fingertip $i$.

On our real-world Shadow Hand, BioTac\textregistered{} sensors \cite{Fishel_2012,BioTac} provide a more complex array of tactile signals. To compute the forces exerted by each finger, readings from the pressure channel of each tactile sensor were acquired and then normalized to match the range of the simulated tactile sensors. In this way, it was possible to directly compare results in simulation and on the real robot, without having to change the parameters of the task or learning algorithm.

The state consists of proprioception (joint position and joint velocity) and touch. The action space is 20-dimensional. We use position control and a control rate of 20 Hz, both in simulation and on the physical robot.

The environment consists of the Shadow Hand and a single block (\figref{fig:gentlemanip}, \figref{fig:setup_realrobot}); the task reward $r_t$ depends on the experiment. Focusing on this simple environment enables us to clearly characterize the effectiveness of our three approaches for training low-impact policies. We find that even in this simple environment, learning policies for gentle manipulation is challenging for most approaches. Results from the simulated environment are presented first.

\subsection{Exploration with impact penalty}
First, we are interested in whether these approaches enable training policies that are gentle during exploration. We investigate this in a no-reward setting, where the policy receives intrinsic rewards (either from dynamics-based surprise or penalty-based surprise) and the intrinsic pain penalty, but no task reward. The goal is for policies to be gentle (i.e., experience low impact) while still exploring effectively, in terms of interacting with objects in the environment.

\begin{figure}[!ht]
\centering
\includegraphics[width=\linewidth]{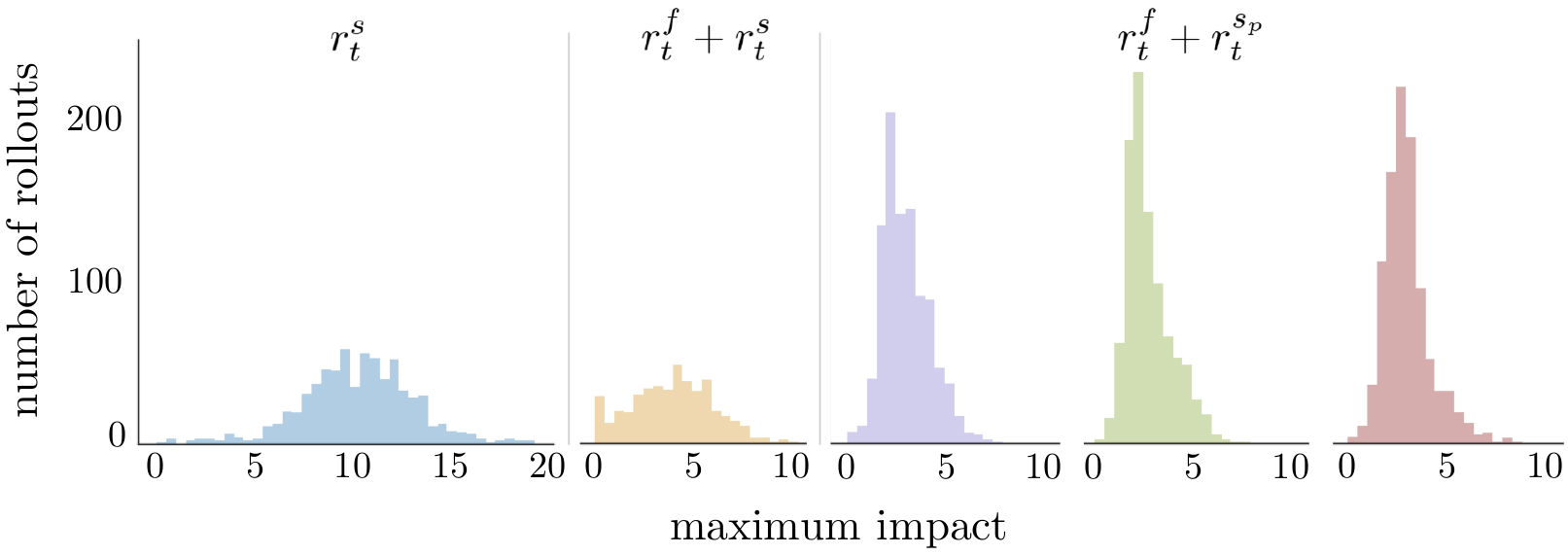}
\caption{We train policies in a no-reward setting, with either dynamics-based surprise ($r_t^s$), left; dynamics-based surprise plus impact penalty ($r_t^f + r_t^s$), center; or penalty-based surprise plus impact penalty ($r_t^f + r_t^{s_p}$), right. The histograms show the maximum amount of impact (in Newtons) per rollout; rollouts are collected regularly throughout 500k training steps, and rollouts with no impact at all are ignored. $\lambda_2' = 2$ \crule[plot-r-rg-rsp1]{0.25cm}{0.25cm}, 3 \crule[plot-r-rg-rsp2]{0.25cm}{0.25cm}, or 4 \crule[plot-r-rg-rsp3]{0.25cm}{0.25cm}.
$\lambda_1' = 2$ for all.}
\label{fig:impact_noreward}
\end{figure}

As a baseline, we trained policies with only dynamics-based surprise intrinsic rewards, without an impact penalty. As expected, these policies experience a large amount of impact while exploring: the maximum amount of impact experienced per rollout is in the 5 to 15N range (\figref{fig:impact_noreward}, left). This suggests that this form of curiosity is not practical for running on real-world robots, if either the robot or the objects it interacts with are susceptible to wear and tear.

When we add the impact penalty, we do observe more gentle exploration: there is a significant decrease in the maximum amount of impact experienced per rollout (now in the 0 to 5N range), for both kinds of intrinsic reward. However, having penalty-based surprise intrinsic rewards leads to more gentle touching, whereas dynamics-based surprise leads to the policy exploring interesting configurations of the hand, but with limited touching (\figref{fig:impact_noreward}, center and right).

\begin{figure}[!ht]
\centering
\includegraphics[width=0.95\linewidth]{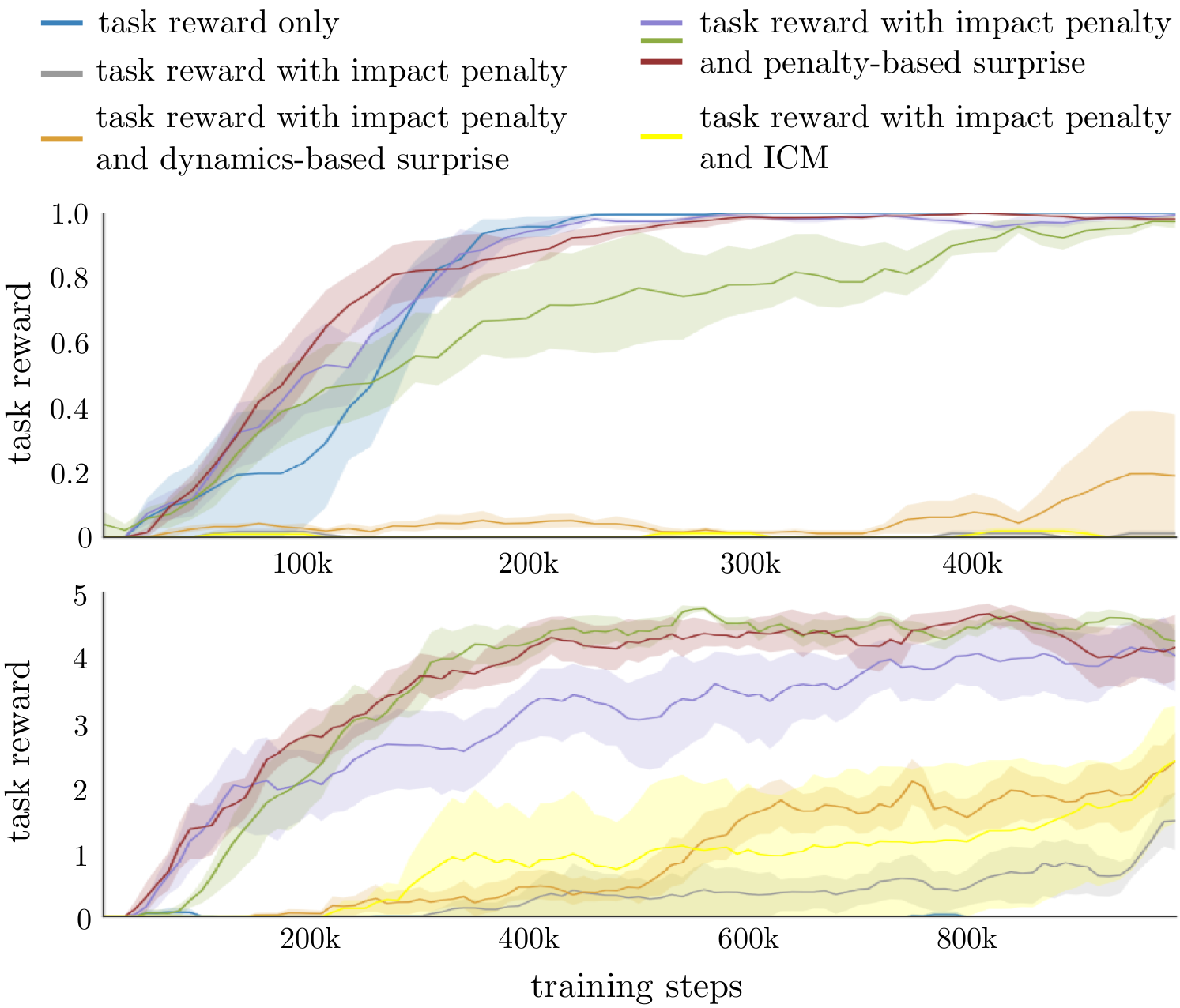}
\caption{
Learning curves for training policies on different reward augmentations (with five random seeds each), for two tasks: pressing a block (\textbf{top}) or a \emph{fragile} block (\textbf{bottom}) with greater than 5N of force. When the block is fragile, the episode terminates with a negative reward if the impact is greater than 3N. Our approach of training policies with a combination of task reward, impact penalty, and penalty-based surprise intrinsic reward is the only one that learns effectively for both tasks. Policies are trained on: task reward only \crule[plot-r]{0.25cm}{0.25cm}; task reward with impact penalty and no intrinsic rewards \crule[plot-r-rg]{0.25cm}{0.25cm}, dynamics-based surprise \crule[plot-r-rg-rs]{0.25cm}{0.25cm}, or penalty-based surprise. The parameterization $\lambda'$ for acceptability of penalties varies: $\lambda_2' = 1.5$ \crule[plot-r-rg-rsp1]{0.25cm}{0.25cm}, 2 \crule[plot-r-rg-rsp2]{0.25cm}{0.25cm}, or 3 \crule[plot-r-rg-rsp3]{0.25cm}{0.25cm} (top) and $\lambda_2' = 1$ \crule[plot-r-rg-rsp1]{0.25cm}{0.25cm}, 1.5 \crule[plot-r-rg-rsp2]{0.25cm}{0.25cm}, or 2 \crule[plot-r-rg-rsp3]{0.25cm}{0.25cm} (bottom). $\lambda_1' = 2$ for all. Policies trained on only task reward are unable to learn the fragile-block task at all (bottom). ICM agents \crule[plot-r-rg-rsp6]{0.25cm}{0.25cm} also do not learn the tasks successfully.
}
\label{fig:learningcurves}
\end{figure}
\begin{figure}[!ht]
\centering
\includegraphics[width=\linewidth]{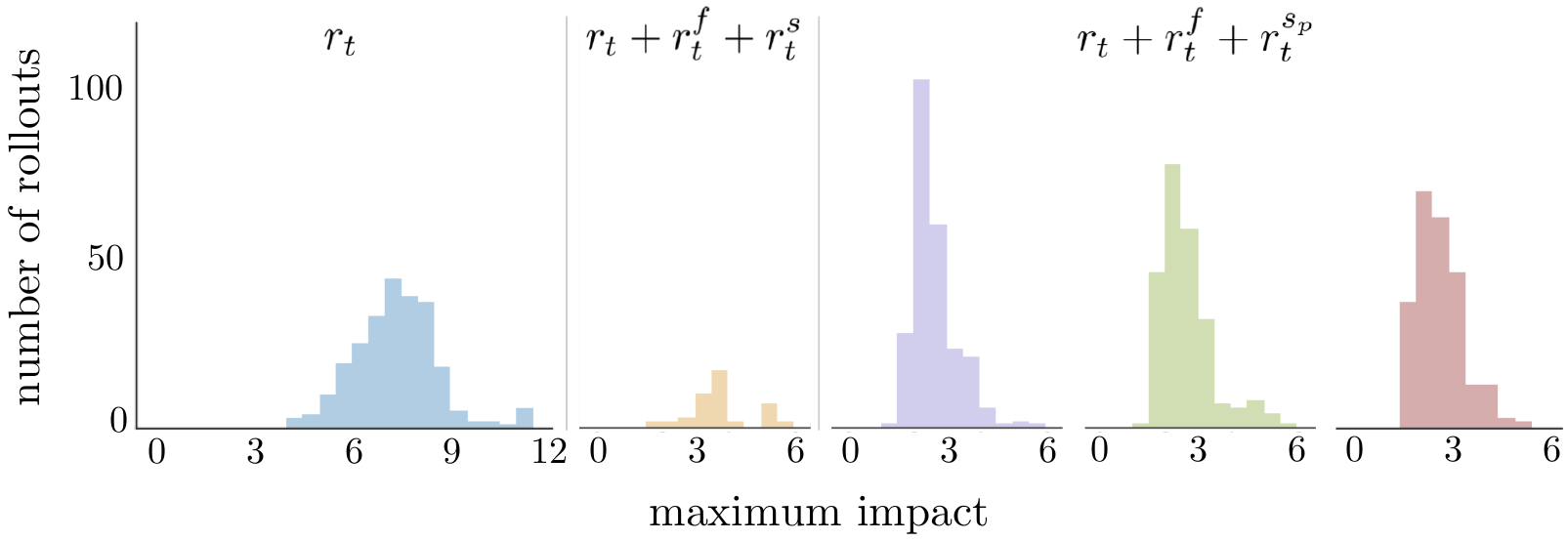}
\caption{We train policies to press a block with greater than 5N of force. These histograms show the maximum impact (in Newtons) experienced per rollout, when the agent performs the task successfully. Rollouts are collected after 500k training steps. $\lambda_2' = 1.5$ \crule[plot-r-rg-rsp1]{0.25cm}{0.25cm}, 2 \crule[plot-r-rg-rsp2]{0.25cm}{0.25cm}, or 3 \crule[plot-r-rg-rsp3]{0.25cm}{0.25cm}. $\lambda_1' = 2$ for all. (Note: with only an impact penalty, agents never succeed in performing the task, so that histogram is not shown.)\vspace*{-0.1cm}}
\label{fig:impact_touchblock}
\end{figure}

\subsection{Manipulation with impact penalty}
Next, we are interested in whether these approaches enable training policies that learn how to perform a task gently, while still being relatively sample-efficient. In the task, the episode terminates with a reward of +1 if the hand presses the block with any fingerpad (thus activating the touch sensor) with a force  greater than 5N. A non-gentle way of achieving this is to go from no contact to 5N of applied force in a single timestep; in contrast, policies trained to be gentle should more gradually increase to 5N of applied force.

This task is simple: without an impact penalty, agents learn this task quickly (\figref{fig:learningcurves}, top), although with a significant amount of impact (\figref{fig:gentlemanip}, top center). However, once impact penalties are added, if there is no form of surprise-based intrinsic rewards to counteract them, then agents fail to learn the task---they get trapped in a local optimum of avoiding contact with the environment, since they experience penalties from contact before discovering how to perform the task (\figref{fig:gentlemanip}, center).

In line with the results from our previous experiment, in which we observed that dynamics-based surprise leads to only limited gentle touching in the presence of impact penalties, we saw that penalty-based surprise was much more effective in terms of agents learning how to perform the task gently, with low impacts (\figref{fig:impact_touchblock}). Even more, these agents learned as quickly as ones trained without the impact penalty (\figref{fig:learningcurves}, top). This may be because this task is particularly contact-focused (in general manipulation tasks are contact-focused, but to varying degrees), so it is a setting in which contact-focused exploration is especially helpful.
We also compare our approach with agents trained using the Intrinsic Curiosity Module (ICM) proposed in \cite{Pathak_2017}. Similarly to the agents trained with dynamics-based surprise, these agents do not successfully learn the task. 

\subsection{Manipulation of fragile objects}
Finally, we made the task more difficult by introducing a `fragile block'. This is analogous to a manipulation task involving fragile objects, such as picking ripe fruit or assisting humans. This fragile block `breaks' if the impact force at any point is greater than 3N, and the episode terminates with a negative reward of -0.5. The reward for completing the task is +5. Now, policies trained with only the task reward are unable to learn the task at all, because they accidentally break the block a few times, and learn that any contact with the block is undesirable. There is no reward shaping that incentivizes these policies to try interacting with the block in a gentle way.

In contrast, policies trained with the impact penalty are better able to learn the task. As before, penalty-based surprise intrinsic rewards are more effective than dynamics-based ones in terms of how quickly policies are able to learn the task (\figref{fig:learningcurves}, bottom).

\begin{figure}[!ht]
\centering
\includegraphics[width=0.95\linewidth]{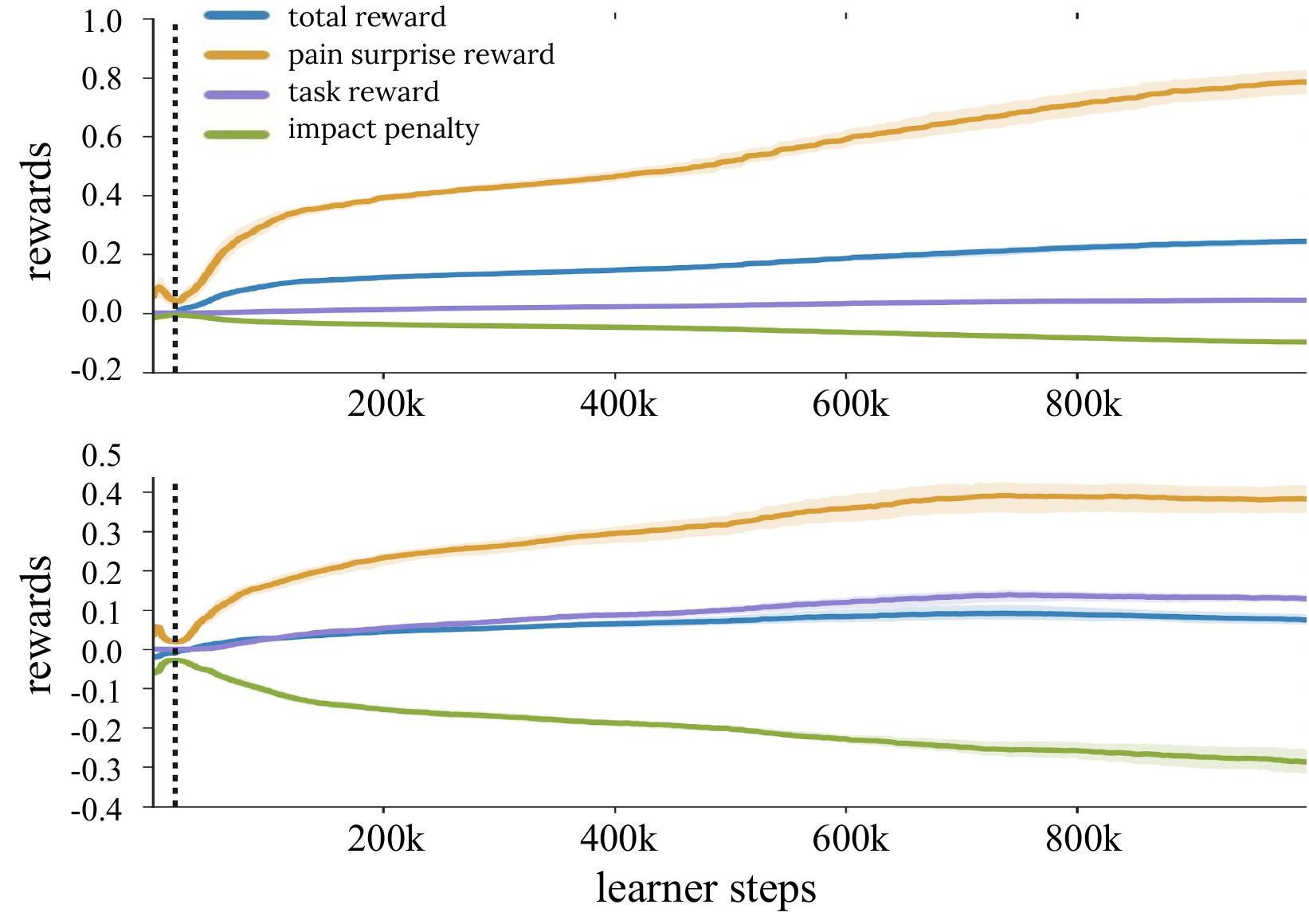}
\caption{Reward components relative to the agent trained on task reward, impact penalty and penalty-based surprise intrinsic reward, for the two tasks: pressing a block (\textbf{top}) or a \textit{fragile} block (\textbf{bottom}). Each point in time represents the average reward (for each component) over a batch of samples. 
}
\label{fig:reward_bonuses}
\end{figure}

\figref{fig:reward_bonuses} plots the evolution of reward components over time obtained by the agent trained on task reward, impact penalty and penalty-based surprise intrinsic reward.
Each reward component in this plot is the average over batches sampled from the agent's replay buffer. The total reward (blue) is the combination of the pain surprise reward, the task reward, and the pain penalty.
The dashed vertical line indicates the timestep when the intrinsic reward starts to be applied (i.e. after 20,000 timesteps). Before this point, the total reward is only affected by the task reward and the impact penalty.

\subsection{Real robot experiments}

\begin{figure}[tp]
\centering
\includegraphics[width=0.7\linewidth, trim={0 2cm 13cm 2cm},clip]{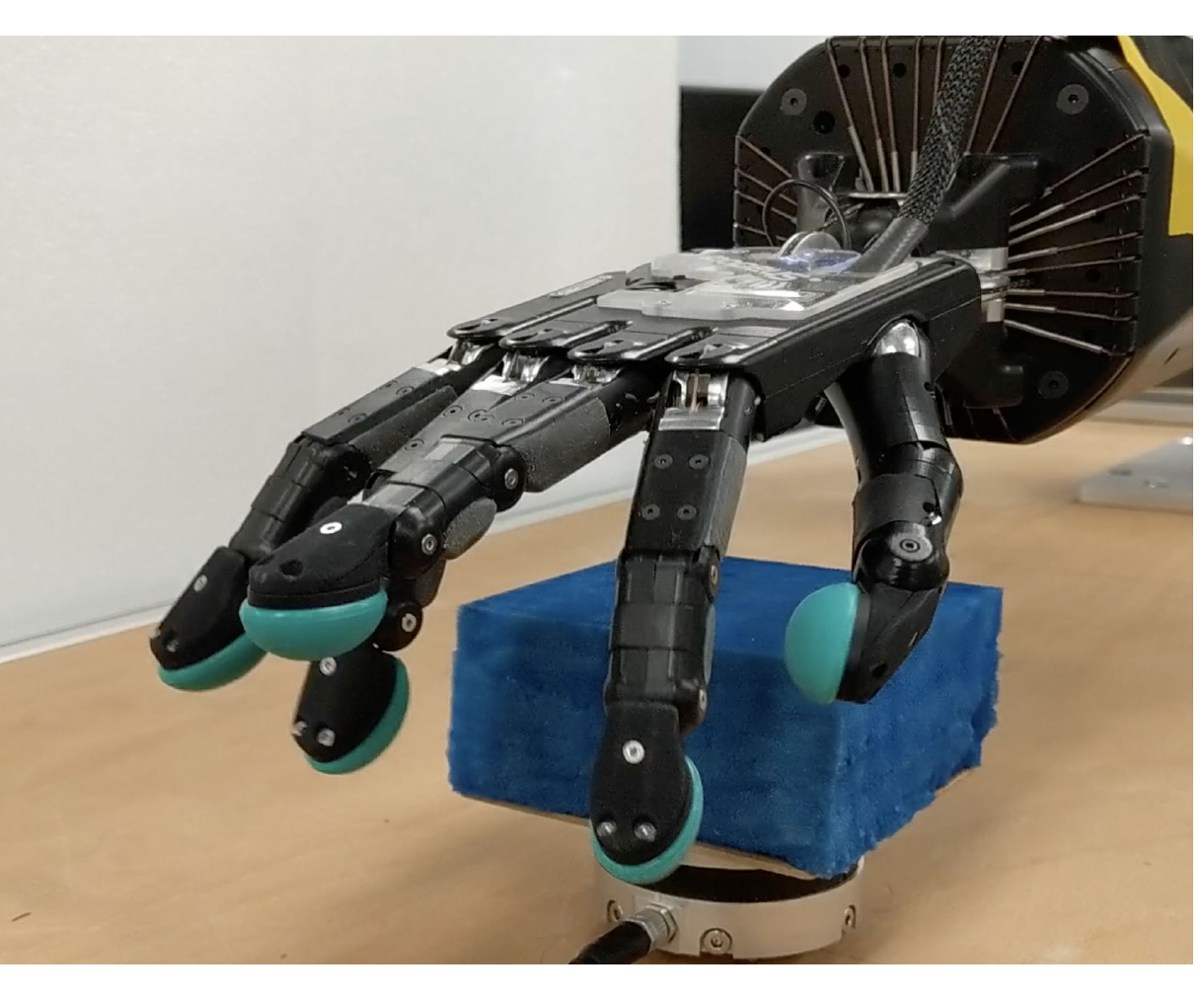}
\caption{Experimental setup on the real Shadow Dexterous Hand \cite{ShadowHand}. A force/torque sensor is attached to a foam block, to measure the force applied to the block. BioTac\textregistered{} tactile sensors \cite{BioTac} were used to compute the forces exerted by the corresponding fingers.}
\label{fig:setup_realrobot}
\end{figure}

We also conducted experiments for the two manipulation tasks on a real Shadow Dexterous Hand \cite{ShadowHand}. The setup used for these experiments is shown in \figref{fig:setup_realrobot}. A force/torque sensor is attached to a foam block, and is used to measure the force on the block.

As in the experiments in simulation, for the first task, the episode terminates with a reward of +1 if the hand presses the block with any fingerpad with a force greater than 5N. For the fragile objects case, the episodes terminate with a negative reward if the impact force at any point is greater than 3N, and the reward for completing the task is +5. 

As baselines, we also ran a random agent, and agents trained using the Intrinsic Curiosity Module (ICM) proposed in \cite{Pathak_2017}. Although these agents sometimes randomly hit the block, they do not consistently solve the tasks, and their interaction with the block often includes high-impact forces that exceed 5N. 

\begin{figure}[tp]
\centering
\includegraphics[width=0.95\linewidth]
{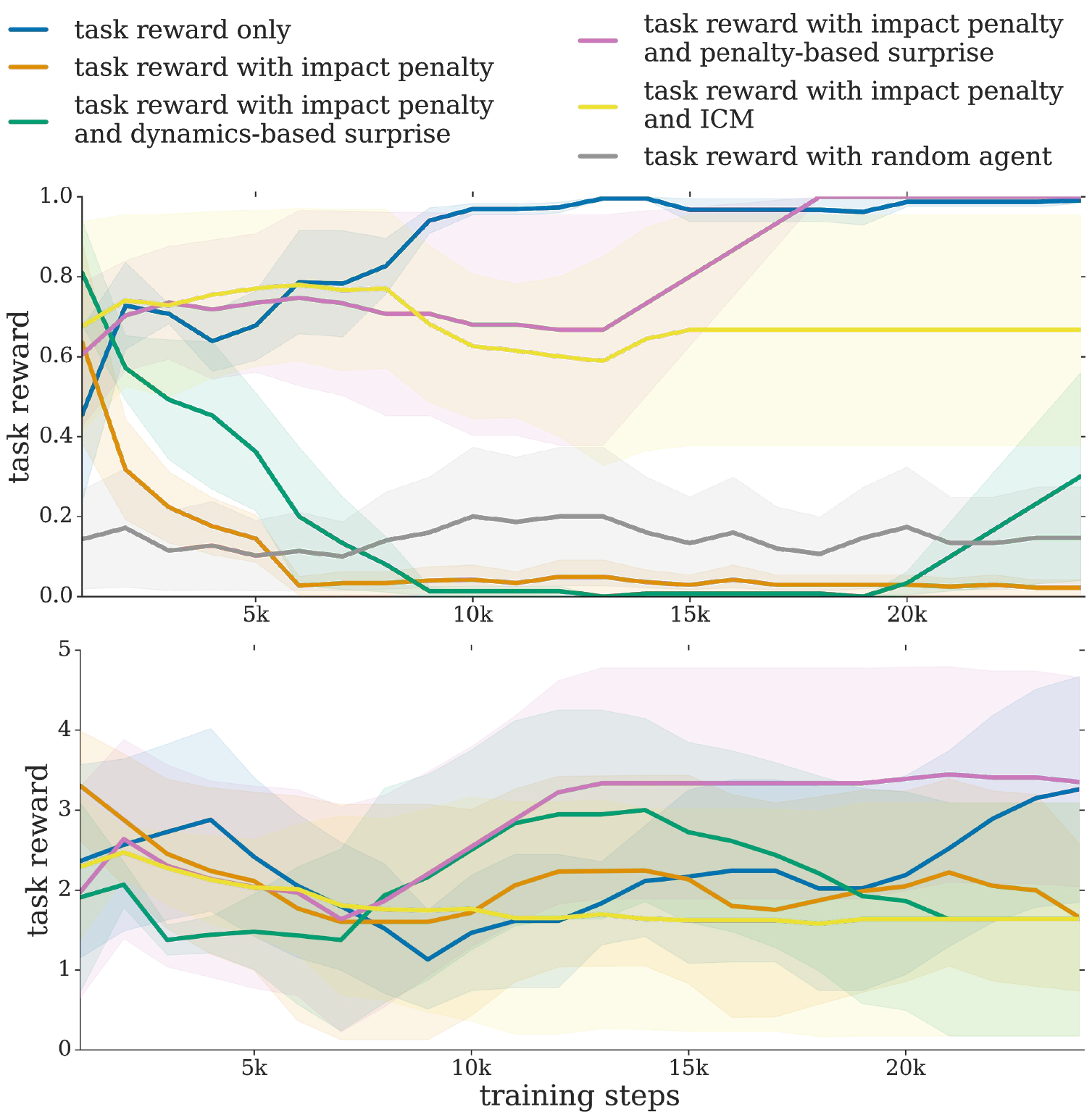}
\caption{Learning curves for training policies on different reward augmentations (with three seeds each), on the real Shadow hand for two tasks: pressing a block (\textbf{top}) or a \emph{fragile} block (\textbf{bottom}) with greater than 5N of force. When the block is fragile, the episode terminates with a negative reward if the impact is greater than 3N. Our approach of training policies with a combination of task reward, impact penalty, and penalty-based surprise intrinsic reward is the only one that learns effectively for both tasks. 
Policies are trained on: 
task reward only \crule[plot-r]{0.25cm}{0.25cm}, 
task reward with impact penalty and no intrinsic rewards \crule[plot-r-rg-rsp7]{0.25cm}{0.25cm}, 
dynamics-based surprise \crule[plot-r-rg-rsp4]{0.25cm}{0.25cm}, or penalty-based surprise \crule[plot-r-rg-rsp5]{0.25cm}{0.25cm}. 
The parameterization $\lambda'$ for acceptability of penalties is $\lambda_1' = 2$ and $\lambda_2' = 1.5$.
As baselines, we also report a random agent \crule[plot-r-rg]{0.25cm}{0.25cm} on the touch task, 
and ICM agents \crule[plot-r-rg-rsp6]{0.25cm}{0.25cm}.
}
\label{fig:learningcurves_realrobot}
\end{figure}

\begin{figure}[pt]
\centering
\includegraphics[width=0.9\linewidth, trim={0.8cm 4.5cm 8cm 4cm},clip]{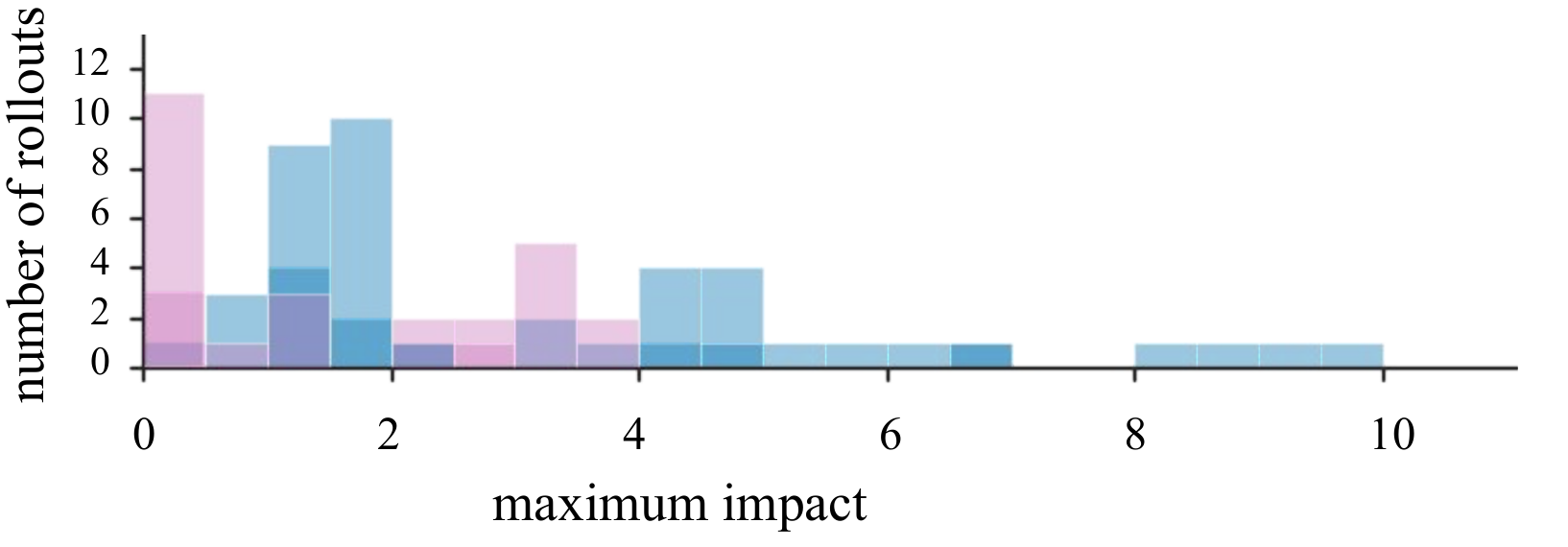}
\caption{We train policies to press a block with greater than 5N of force. These histograms show the maximum impact (in Newtons) experienced per rollout, when the agent performs the task successfully. 
Colors refer to agents trained on task reward only \crule[plot-r]{0.25cm}{0.25cm} and penalty-based surprise \crule[plot-r-rg-rsp5]{0.25cm}{0.25cm}. 
$\lambda_1' = 2$ and $\lambda_2' = 1.5$.
(Note that agents trained on task reward with impact penalty and no intrinsic rewards and dynamics-based surprise, the random agent and the ICM agents don't learn the task, hence the corresponding histograms are not shown.)}
\label{fig:impact_touchblock_realrobot}
\end{figure}

We evaluated the different agents over 25,000 training steps. Since this is a shorter time window compared to that of the experiments executed in simulation, we increased the learning rate by one order of magnitude (from 0.0001 to 0.001). In line with the results obtained in simulation,  we saw that penalty-based surprise was much more effective in learning how to perform the task gently, both in terms of learning speed (\figref{fig:learningcurves_realrobot}) and minimizing impacts (\figref{fig:impact_touchblock_realrobot}).
The agent trained with dynamics-based surprise continues exploring the (complex) dynamics of the system on the real robot, and thus struggles to learn the simple manipulation task.
On the other hand, the agent trained with penalty-based surprise is not only successful on the simple manipulation task, but notably it is the one that learns the task of manipulation of fragile objects more consistently.

%% file: discussion.tex
\section{Discussion and Future Work}
Our work takes a step toward using deep RL to train policies for gentle, contact-rich manipulation. Although curiosity has long been established as a source of endogenous motivation for artificial agents exploring the world, it may be too broad and general to drive an agent towards contact-rich policies, especially when penalties are used to discourage high-impact interactions. We found that in this scenario, choosing the appropriate focus of curiosity is important for incentivizing agents to interact gently with the environment. This enables efficient and safe exploration, precise task execution, and successful manipulation of fragile objects. Although the proposed approach is demonstrated on relatively simple tasks, we believe that it paves the way towards a new direction in curiosity research, one that identifies more nuanced types of curiosity and intrinsic motivation for deep RL agents. 

A main direction of future work is to apply this approach to more complex tasks, for instance in dynamic environments. In addition, this work only considers one aspect of being gentle---impact. Our approach could be used to train policies while minimizing other sources of wear and tear, for instance total force (rather than the increase in force), or the torques exerted by a robot's motors (which would reduce energy consumption as well \cite{Mohammed_2014}). Additionally, we note that although we use a multimodal robot environment---integrating tactile and proprioceptive sensors---we have not incorporated vision, which would provide an additional observation to support tactile and force predictions. Future work will seek to establish the value of contact-focused curiosity across this broader multimodal landscape.